# Vision-based localization methods under GPS-denied conditions

Zihao Lu, Fei Liu, Xianke Lin*

Department of Automotive and Mechatronics Engineering, Ontario Tech University, Oshawa, ON L1G 0C5, Canada;

*Correspondence: xiankelin@ieee.org; Tel.: +1-905.721.8668 ext. 2819

**Abstract:**
This paper reviews vision-based localization methods in GPS-denied environments and classifies the mainstream methods into Relative Vision Localization (RVL) and Absolute Vision Localization (AVL). For RVL, we discuss the broad application of optical flow in feature extraction-based Visual Odometry (VO) solutions and introduce advanced optical flow estimation methods. For AVL, we review recent advances in Visual Simultaneous Localization and Mapping (VSLAM) techniques, from optimization-based methods to Extended Kalman Filter (EKF) based methods. We also introduce the application of offline map registration and lane vision detection schemes to achieve Absolute Visual Localization. This paper compares the performance and applications of mainstream methods for visual localization and provides suggestions for future studies.

# 1. Introduction

Autonomous navigation systems need to complete tasks such as motion planning, path tracking, obstacle avoidance, and target detection, which require the system to have the ability to estimate, perceive and understand the environment [1]. Global Navigation Satellite Systems (GNSS) provide reliable environmental information and real-time positioning for Unmanned Aerial Vehicles (UAVs) and Autonomous Vehicles (AVs). But if UAVs perform missions in challenging and cluttered environments, GNSS signals could be lost or suffer from fading, multipath effects, jamming and spoofing. So it will not work properly when a UAV is flying at a low altitude in terrain full of obstacles such as cities, canyons/mountains, or forests [2]. To address this issue, in recent years, researchers have developed a series of localization solutions for UAVs, such as lidar-based and vision-based methods. In robotic localization research, using multiple sensors is very costly, and having too many devices consumes additional power and brings additional weight. These problems have hindered the commercialization of various robot localization algorithms [1]. Visual localization is an attractive alternative, which is a computer vision-based technology. Its working principle is to capture images of the surrounding environment through a visual camera and then calculate the location and direction of the surrounding environment, i.e., mapping the unknown environment. Its advantages include: the camera cost is relatively low, and it can obtain rich environmental information, including visual information such as color and texture. However, visual localization requires high computing power, images take up a lot of storage space, and software development is also relatively difficult. In addition, the visual system is greatly affected by light, and it is not easy to function in a poorly lit environment.

Currently, vision-based localization methods include two main methods: Relative Visual Localization (RVL) and Absolute Visual Location (AVL). RVL includes Visual Odometry (VO) and Visual Simultaneous Localization and Mapping (VSLAM). Studies on AVL mainly include closed-loop VSLAM [3–11], content-based remote sensing image retrieval [12, 13], registration of UAV images to satellite images [14–16], location recognition, and image geolocation [17–25]. VO is a method to estimate agent ego-motion using monocular or binocular cameras. The algorithm mainly focuses on the local consistency of robot trajectory, estimating the pose, and performing local optimization at the same time. VSLAM is a method to build a map of the environment with the help of a camera without any prior information about the surrounding environment and obtains an estimate of global consistency on robot trajectory [26]. RVL often requires knowing its absolute position at the beginning and using a vision system in combination with other sensors to estimate its current position and current velocity. Therefore, the problem of RVL is the accumulation of estimation errors over time. Currently, there are many studies aimed at addressing the drift of VO in state estimation [27–29]. For VSLAM, since the method obtains a globally consistent map estimation, it can also effectively reduce the drift in the state estimation by revisiting the previously mapped regions. In contrast, AVL mainly relies on offline map registration, so AVL often does not have drift problems in position and state estimation, and it has a known error range and estimation accuracy. Therefore, AVL can obtain the absolute position by matching the edge features of the environmental image and the geo-referenced image [30].

This paper summarizes the research status and challenges of localization technologies in GPS-denied environments and introduces the latest algorithms in this field. Finally, we present the research prospects of localization technologies. The main contributions are summarized in the following four aspects.

1) The working principle of optical flow and its algorithm are introduced, and the wide application of optical flow in visual odometry is introduced. This paper focuses on the introduction of FlowNet and its subsequent improved algorithms and makes a comparison to guide the selection of optical flow estimation algorithms based on machine learning.

2) Classifying the latest research progress of visual SLAM algorithms in localization and describing their respective characteristics.

3) The potential of image registration in localization is comprehensively evaluated, and localization methods based on offline map registration and lane detection are introduced.

4) Summarizing the main challenges in the development of localization technologies under GPS-denied conditions and potential solutions. The research prospects in this field are introduced which helps researchers to focus on the main research problems and bottlenecks.

This review article begins with an introduction to the application of optical flow techniques in visual odometry and discusses the contributions of visual odometry, visual SLAM, and image registration techniques to localization. Following the discussion of state-of-the-art technologies, we compare the advantages and disadvantages of different technologies and present the future research directions of localization technologies under GPS denial conditions. The remainder of this review paper is organized as follows. Section 2 introduces optical flow-based visual odometry techniques and presents a broad range of visual odometry applications. Then, Section 3 introduces the basic working principles of visual SLAM technology and introduces the latest VSLAM frameworks and their applications. Section 4 describes the application of image registration technology (including offline map registration and lane detection) in localization. In Section 5, we present some key challenges in this field and prospects for future research. Finally, Section 6 gives a brief conclusion.

## 2. Relative Visual Localization (optical flow-based)

VO is a vision-based navigation method that estimates the motion of the robots (rotation and translation) and localizes itself in the environment by comparing the differences between frames. Many VO methods have been proposed in recent years, and these methods can be divided into monocular camera methods and stereo camera methods according to the type of camera used. The binocular camera is the most commonly used stereo camera, which can use the distance between the two cameras to obtain depth information. RGBD cameras can acquire image and depth information at the same time, but the acquired depth range is limited, which is greatly affected by infrared light and consumes a lot of power. Monocular cameras are simple in structure and low in price, and many studies are currently carried out on monocular cameras. The most widely used vision sensor in the VO method is the monocular camera. Unlike other sensors, this sensor is affordable, compact, and energy efficient. Therefore, they can be easily installed on small platforms such as UAVs [31, 32].

VO can also be divided into the direct method and feature-based method according to whether it is necessary to extract features. The feature-based method is considered to be the mainstream method of VO because of its advantages of strong rotation robustness and fuzzy robustness and strong scale transformation. The main working principle of this method is to estimate the motion pose of the camera by selecting representative points (such as corner points) in the image and simultaneously analyzing the motion state of the corresponding feature points in the two frames before and after. At present, many mature feature extraction methods have been proposed in the field of computer vision, such as SIFT [33] (construction of features based on the histogram of gradient direction and gradient value) , SURF [34] (construction of histogram based on the magnitude of gradient value) and ORB [35] (size based on pixel value) . The following figure is a schematic diagram of Chen et al. using the SURF algorithm to detect and match feature points. They use the Approximate Nearest Neighbor (ANN) algorithm for matching. First, the SURF features extracted from all template flags are divided into eight groups and stored, and then the features of the images to be processed are compared with the database respectively to match the candidate images, and finally the features are matched using ANN [36].

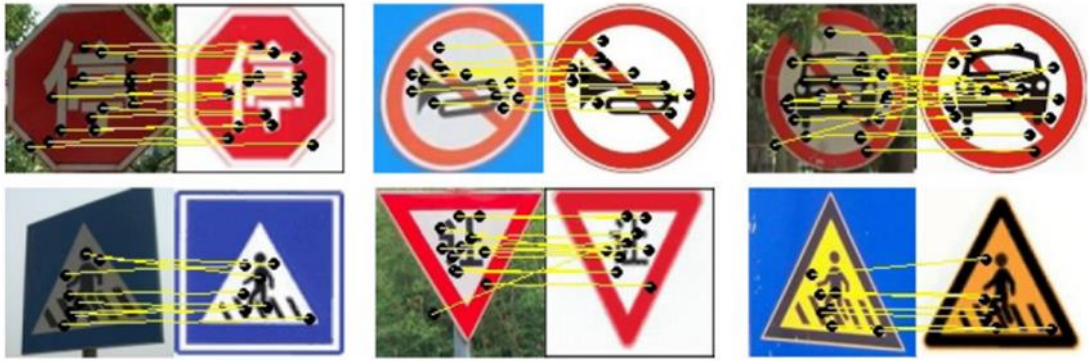

Fig.1 The SURF algorithm extracts feature points and matches them. The left side of each subgraph is the image to be processed, and the right side is the image in the database. The number of matching points is 16, 11, 24, 7, 12, and 7, respectively. [36]

In the research work by Zheng et al. [37], they used different algorithms for feature extraction on images in five different scenes. The results in the table below show that SIFT takes the longest time to compute, while ORB does the computation faster .

Table.1 Point and Time comparison of three feature extraction algorithms (SIFT, SURF, and ORB) [37]

| *No.* | *SIFT* | | *SURF* | | *ORB* | |
|---|---|---|---|---|---|---|
| | *Point* | *Time(s)* | *Point* | *Time(s)* | *Point* | *Time(s)* |
| 1 | 171 | 0.01889 | 86 | 0.01244 | 168 | 0.00282 |
| 2 | 253 | 0.01899 | 254 | 0.01621 | 299 | 0.00411 |
| 3 | 234 | 0.01966 | 187 | 0.01386 | 251 | 0.00470 |
| 4 | 175 | 0.02557 | 183 | 0.01570 | 168 | 0.00221 |
| 5 | 24 | 0.01669 | 20 | 0.01038 | 19 | 0.00175 |

Although the feature-based method is the mainstream approach in VO, this method still has shortcomings. For example, it requires high computing power, and the extraction of features and the calculation of descriptors are very time-consuming. At the same time, since this method generally only extracts a few hundred features to represent image motion, a lot of information is inevitably lost relative to the hundreds of thousands of pixels in an image. This problem is especially obvious when the image texture is insufficient. Since the features that can be extracted from the image are greatly reduced, it is difficult for the feature-based method to estimate the camera motion in this case correctly. Therefore, researchers are trying to use optical flow to improve the feature-based method.

Optical flow techniques are inspired by the localization mechanism of birds and insects during flight and are used to solve navigation problems [38]. At present, this technology has been widely used in robot positioning and navigation to provide reliable speed and position information. Optical flow method uses the changes of pixels in the image sequence in the time domain and the correlation between adjacent frames to find the corresponding relationship between the previous frame and the current frame, to calculate the motion of objects between adjacent frames. Usually, the instantaneous change rate of gray level on a specific coordinate point of the two-dimensional image plane is defined as the optical flow vector. The advantage of the optical flow method is that it can accurately detect and identify the position of the moving target without knowing the information of the scene, and it is still applicable when the camera is in motion. Optical flow not only provides information about the unknown environment, but also helps determine the direction and speed of the robot, and can detect moving objects without knowing any information about the scene. At the same time, due to the relatively mature development of optical sensors, their cost is usually low and it is easy to be miniaturized, which can effectively reduce costs and increase portability [27].

## 2.1 Optical Flow Calculation

Optical flow is the instantaneous velocity of the pixel of moving objects on the imaging plane and can be viewed as the apparent motion of objects, brightness patterns, or features observed from the eye or camera [39]. Optical flow can be computed in a dense manner [40] or a sparse manner [41]. The dense optical flow algorithm calculates the motion vector of all the pixels of the two frames before and after, while the sparse optical flow algorithm only calculates the feature points in the image. The dense optical flow has high accuracy, but it has the disadvantage of requiring a lot of computational power to perform calculations on all image pixels. In contrast, sparse optical flow only needs to calculate the optical flow vector of a few features in the whole image. Although it is not as accurate as dense optical flow, its calculation time is shorter.

In the feature-based method based on optical flow, the matching descriptor of the original feature (used to describe the moving distance and direction of the feature) is replaced by the optical flow vector. Since the computation time of the optical flow vector is less than that of the computational descriptor, the algorithm efficiency can be greatly improved. Similarly, direct methods developed from optical flow methods can directly estimate camera motion by minimizing the photometric error (i.e., minimizing the reprojection error of features in the feature-based method) without extracting features or computing feature descriptors [42]. The direct method avoids both the calculation of features and the missing of features.

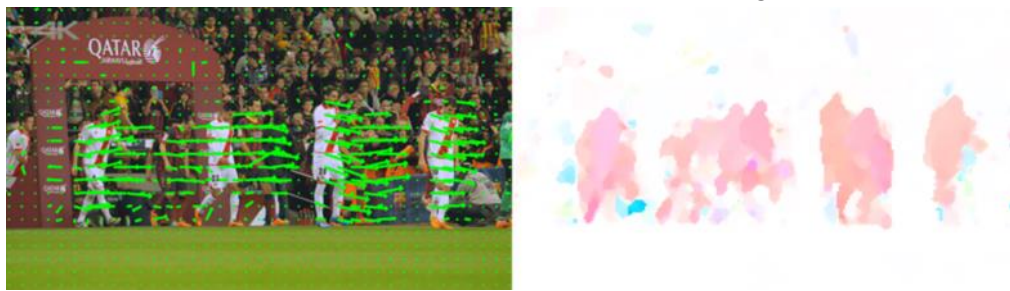

Fig.2 The left picture is a schematic diagram of the optical flow field, and a relatively drastic optical flow change can be seen. The right picture is the moving object segmented according to the optical flow vector threshold. [43]

In the following section, we will classify and describe the traditional methods and deep learning methods in the recent research on optical flow estimation methods.

2.1.1 Traditional Methods of Optical Flow Calculation

At present, researchers have proposed various schemes for calculating optical flow, such as Lucas-Kanade algorithm [41], HornSchunck algorithm [44], image interpolation algorithm [45], block matching algorithm [46], feature matching algorithm [33], etc. Optical flow estimation is generally based on the assumption of constant brightness and smoothness. Constant brightness assumes that the apparent brightness of an object remains constant between two consecutive frames. The constant smoothness assumes that the movement is small, i.e., pixel values are similar in the neighborhood [44].

In actual situations, lighting conditions generally do not satisfy the assumption of constant brightness of adjacent frames in the optical flow method, that is, changes in lighting conditions will affect the accuracy of optical flow calculations. To address the problem of optical flow calculation under non-constant lighting conditions, Zhang et al. studied an optical flow localization method based on ROF denoising. They use the convex optimization theory and the duality principle to decompose the image under changing lighting conditions and reduce its impact. The decomposed texture is used to calculate the optical flow of the image and further extract the motion information of the position and attitude of the MAV [47]. The constant smoothness assumption is also difficult to satisfy strictly in practice. Boretti et al. utilized the Lucas-Kanade method to compute sparse optical flow fields and detect features with an ORB detector. They adopted an image pyramid approach to avoid the failure of feature detection in large-motion situations [48]. Similarly, Lou et al. [49] applied image texture decomposition and image pyramid techniques to the Lucas-Kanade optical flow algorithm, which reduced the interference of illumination changes and large displacements on moving object detection. The following figure is the comparison between their algorithm and the classic Lucas-Kanade algorithm. It can be seen that the improved algorithm can focus more on the detection of moving objects.

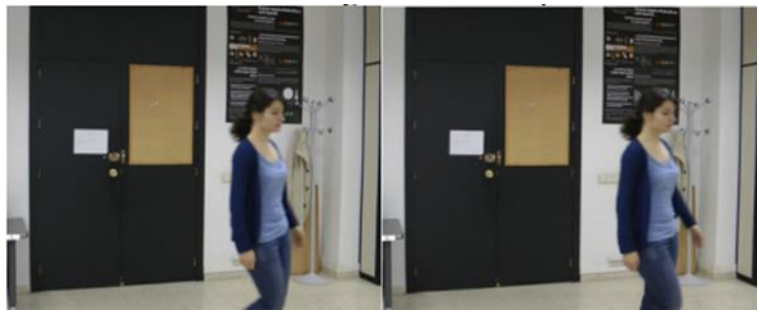

(a) Input image

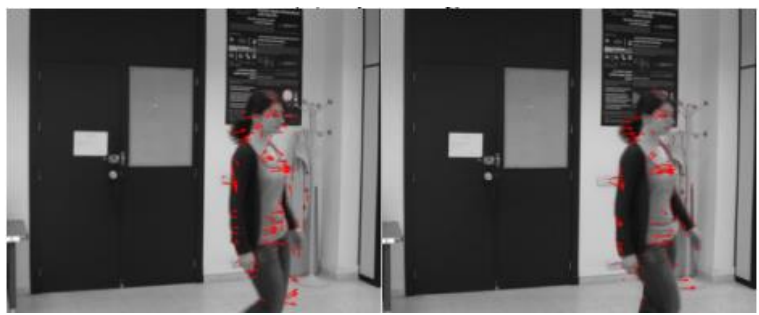

(b) L-K optical flow algorithm

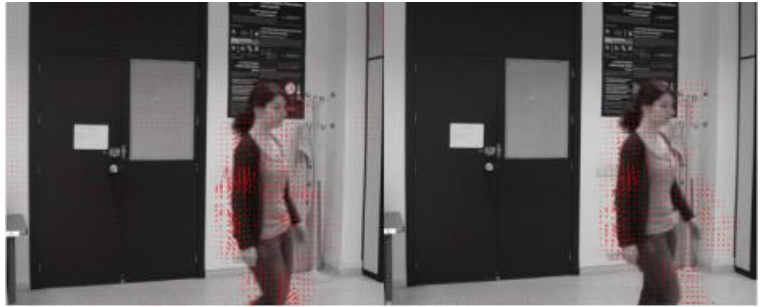

(c) The improved algorithm by Lou etc.

Fig.3 Results of moving object detection [49]

2.1.2 Deep Learning Methods of Optical Flow Calculation

Compared with traditional image-based methods, deep learning methods for estimating optical flow achieve a new level of performance while avoiding explicit modeling of the entire problem, and have strong potential in the field of optical flow estimation [50].

The learning-based optical flow estimation method was first developed by Dosovitskiy et al. [51], who proposed FlowNet to solve the optical flow estimation problem with a supervised learning method. Dosovitskiy et al. [51] also proposed a dataset called Flying chairs and used this virtual synthetic dataset to train the FlowNet network. Experiments show that the FlowNet model under this training set can generalize well to real-world images .

The following figure shows the structure of two FlowNets. The first one superimposes two consecutive frames of input images and passes them through a series of networks with only convolutional layers, called FlowNetSimple. The second one processes two frames of pictures separately. The images are inputted to the convolution layer and their respective features are extracted, and then matching is performaed, which is called FlowNetCorr.

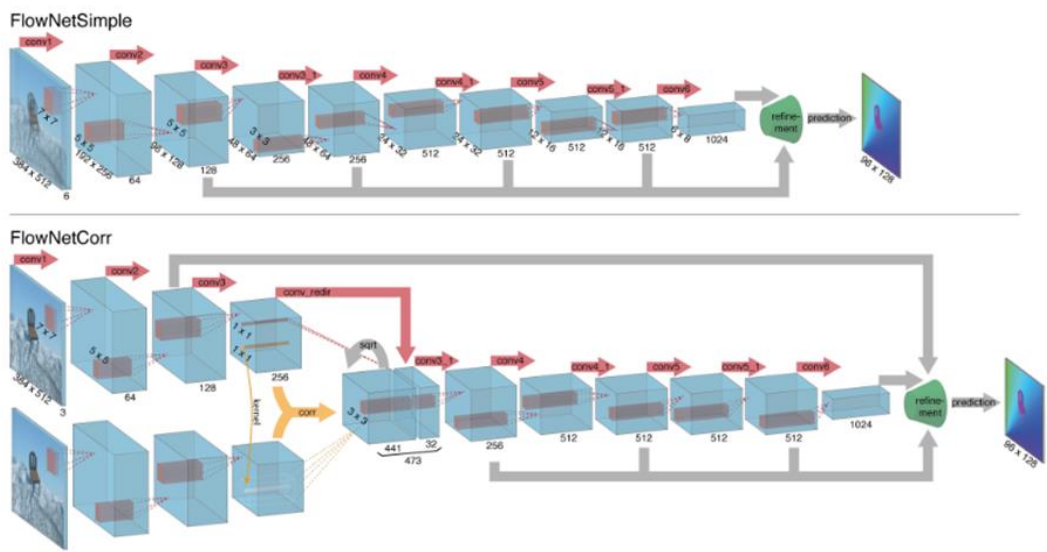

Fig.4 The network structure of FlowNetSimple (top) and FlowNetCorr (bottom), FlowNetSimple processes two frames superimposed, FlowNetCorr processes two frames separately, and the training of the network is an end-to-end process. [51]

However, the disadvantage of FlowNet is that its prediction error rate is too high, and it can not correctly handle small displacements and real-world data. FlowNet2 is an improved version of FlowNet. Llg et al. significantly reduced the estimation error of FlowNet by using a serial multiple network architecture and introduced a branch network to deal with the optical flow estimation problem of small object displacement [52]. Sun et al. made a further improvement and designed PWC-Net. The network uses CNN convolution to obtain image features, and then predicts the optical flow at a higher resolution according to the optical flow estimation at a low resolution according to the pyramid processing principle, and gradually obtains the optical flow at the final resolution [53]. The advantage of PWC Net over FlowNet2 is that it is easier to train. Recently, Zhu et al. [50] proposed EV-FlowNet based on FlowNet, which is a novel self-supervised deep learning channel for event-based camera optical flow estimation. They first used a new method to represent the event flow in the form of images. An image with four channels is applied to the deep learning network, where the first two channels encode positive events and negative events respectively. This event counting is a common method for visualizing the event flow and has been proven to provide information in a learning-based framework to regress the 6-DOF pose [54]. The last two channels of the image are used to encode the timestamp of the last event that occurred on each pixel, to avoid losing the motion rule in the image.

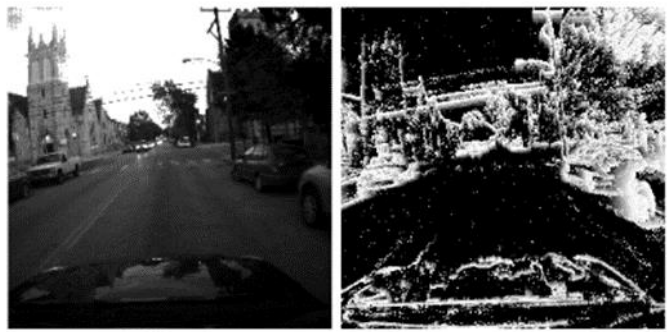

Fig.5 Example with time stamp image. The bright area indicates the occurrence of the event set. [50]

EV-FlowNet takes the image based on a given event stream as the only input, with a size of 256*256*4. Given the estimated traffic from the network, the corresponding gray level image captured from the same camera at the same time with the event is then used as a supervision signal to provide a loss function during training. The combination of images and self-monitoring loss is enough for the network to learn to predict accurate optical flow only from events [50].

The following figure shows the structure of EV-FlowNet. The green area is the convolutional layer, which completes downsampling (encoding), and the convolutional results of each layer are retained and linked to the upsampling (decoding) layer as a skip layer. The middle blue part is the residual block, which can further extract features. The last yellow area is the upsampling (decoding) part, which is achieved by symmetric padding. In the upsampling stage, the results of each layer are convolved and the loss is calculated, and then merged into the upsampling stage to continue the calculation.

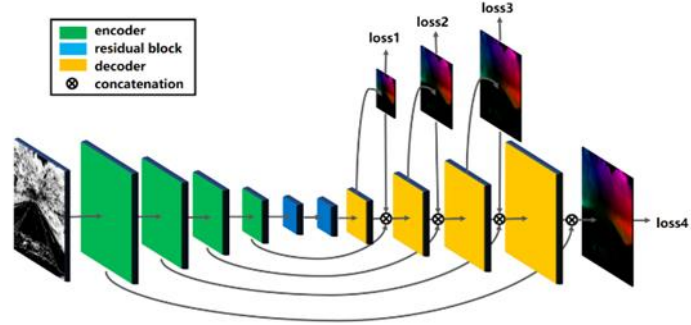

Fig.6 Network structure of EV-FlowNet. The green area is the convolutional layer, which completes downsampling (encoding), and the convolutional results of each layer are retained, which are linked to the upsampling (decoding) layer as a skip layer. The middle blue part is the residual block, which can further extract features. The last yellow area is the upsampling (decoding) part, which is achieved by symmetric padding. Furthermore, each set of decoder activations is passed through another depthwise convolutional layer to generate stream predictions at its resolution. The loss is applied to this stream prediction, and the prediction is also connected to the decoder activation. [50]

Yin and Shi [55] proposed GeoNet, which is a joint unsupervised learning framework compared to EV-FlowNet for monocular depth, optical flow, and ego-motion estimation from videos. The motion information captured by the camera is composed of rigid flow (static features) and non-rigid flow (dynamic features), which are derived from the motion of the camera itself and the motion of the target object. Based on this, they designed a novel cascaded architecture consisting of two stages to calculate the global displacement of the picture and the fine displacement of the picture, respectively, to adaptively solve the estimation problem of rigid and non-rigid flow. The following table shows a comparison of FlowNet and its improved algorithms.

Table.2 Comparison of contributions and disadvantages of FlowNet and its subsequent improved algorithms

| Method | Main Contribution | Main Disadvantages | Dataset |
| --- | --- | --- | --- |
| FlowNet[51] | Created a precedent for CNN to predict the optical flow | Low prediction accuracy and cannot effectively handle small displacement or real-world data | Middlebury[56]+KITTI[57]+Sintel[58]+Flying Chairs[51] |
| FlowNet2[52] | Compared with FlowNet, the accuracy and speed of predicting optical flow are greatly improved | Image noise still has a large impact on prediction | Middlebury+KITTI+Sintel+Flying Chairs |
| PWC-Net[53] | Greatly reduces the size of the CNN network, making it easier to train | PWC-Net upsamples the results, resulting in blurry estimates | KITTI+Sintel+Flying Chairs |
| EV- | Use event camera data | The image recording method | MVSEC[59] |
| FlowNet[50] | as network input | is relatively special and leads to poor migration | |
| GeoNet[55] | Implementing an adaptive solution to the estimation problem of rigid and non-rigid flows | Gradient Locality of Warping Loss, causes GeoNet to perform worse than direct unsupervised flow network. | KITTI+Cityscapes[60] |

Constrained by the limited memory and computing power of embedded processors on low-end robots and UAVs, MuMuni et al. designed a compact CNN suitable for real-time deployment in UAV systems equipped with NVIDIA Jetson TX2 processors [61]. They decomposed UAV navigation into three tasks and used a separate convolutional neural network for each task. In this neural network architecture, two consecutive frames are taken as an input once, DepthNet is used for monocular depth measurement, EgoMNet is used for ego-motion estimation, OFNet is used for optical flow estimation, and ground plane segmentation (G-Seg) map is used for calibration metric scale. In the second stage, masks are computed to filter out appropriate regions and used together with the consistency loss to update the network.

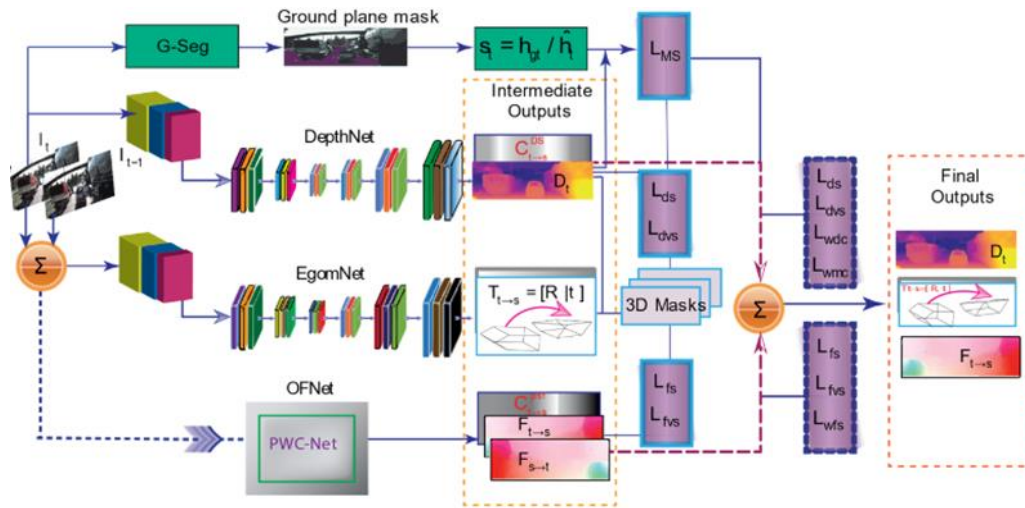

Figure 7: Schematic illustration of the UAV autonomous navigation network designed by MuMuni et al. Ground plane segmentation (G-Seg) maps are also used to calibrate the metric scale. DepthNet estimates depth per frame, EgoMNet estimates relative camera pose, and OFNet predicts optical flow. The network also estimates confidence maps for each task. [61]

## 2.2 Application of VO

VO has received increasing attention due to its wide application in robotics, autonomous driving, and Augmented Reality (AR). Under GPS-denied conditions, VO is widely used due to its low cost and easy access as an effective complement to other sensors such as Inertial Navigation Systems (INS) and wheel odometers. VO methods can be roughly divided into three categories: geometry-based methods, deep learning-based methods, and hybrid methods. Traditional monocular VO methods usually consist of tracking, optimization, and closed-loop modules, which take full advantage of geometric constraints and often use optical flow methods to extract image features. Even though traditional methods often show more robustness and accuracy in pose estimation and navigation, they often suffer from scale ambiguity without additional information. At the same time, deep learning-based methods solve the above problems by training CNNs with large amounts of data. Instead of setting the geometric constraints manually, DL-based methods can obtain them by exploiting prior knowledge in the training data. Even when the parallax is not large enough, reasonable pose and depth can be estimated. Furthermore, online learning can be leveraged to further improve performance. Despite the advantages of DL-based methods, the accuracy of estimating ego-motion is still inferior to traditional methods [62].

This section mainly introduces the application of VO based on optical flow in the direction of multi-sensor fusion deviation correction, estimation of velocity distance and position, obstacle detection, etc.

2.2.1 Multi-sensor fusion

Under the GPS denial condition, the available localization methods are generally: SLAM, inertial IMU localization, and visual localization. SLAM is more accurate but large in size and high in cost; IMU devices often drift, resulting in integral errors; and visual localization requires strong real-time computing power support. At present, the multi-sensor fusion localization scheme has made great progress in reducing localization errors, improving system robustness, and reducing costs. Therefore, integrated localization schemes are being widely adopted by researchers.

Shen et al. [28] proposed a multi-sensor fusion localization algorithm based on Extended Kalman Filter (EKF). Among them, VO is used to measure the speed and position of the UAV, the magnetometer is used to measure the attitude of the UAV, and then the drift of the INS is calibrated using the extended Kalman filter [28]. Kim et al. [63] proposed an improvement to Shen's work by designing a Feature Point Threshold Filter (FPTF) algorithm that can improve the INS+Optical Flow sensor fusion performance by changing the threshold according to the altitude and speed of the UAV.

Since monocular visual odometry cannot accurately measure depth and distance, scale blur is a fundamental problem of monocular VO. Yu et al. [64] try to mine prior knowledge from the environment to solve this problem. In their work, assuming a constant camera height above the ground and fitting the ground plane with the least squares method, an efficient ground point extraction (GPE) algorithm based on Delaunay triangulation [65] and ground points for aggregating ground points from consecutive frames is used. Point Aggregation (GPA) algorithm, which finally calculates the true scale by estimating the ground plane. Based on the aggregated data, the scale is finally restored by solving a least squares problem using a RANSAC-based optimizer. The following figure shows the work of Yu et al.'s algorithm in filtering and selecting ground feature points. The results show that the GPE algorithm can effectively remove outliers and improve tracking accuracy.

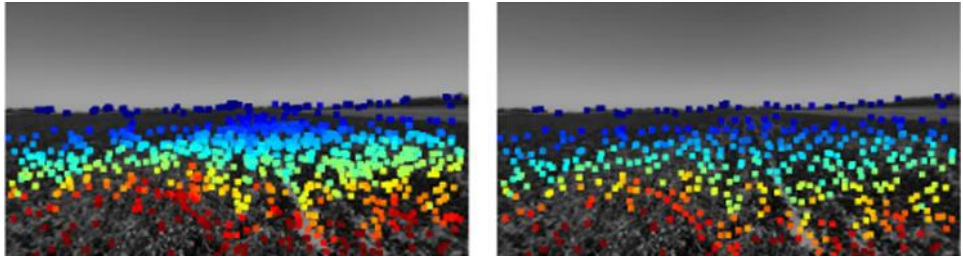

Fig.8 GPE algorithm filters ground feature points Left: Original feature point Right: Filtered feature point [64]

Experiments on the KITTI dataset show that the framework proposed by Yu et al. has competitive performance in terms of translation error and rotation error. Their method exhibits high computational efficiency with a high-frequency performance of 20 Hz on the KITTI dataset [64]. A novel intelligent hybrid vision-aided inertial navigation system was proposed by Mostafa et al. to address the scale ambiguity problem of optical flow in vehicle motion estimation [27]. The system consists of three main modules: VO, Gaussian regression process (GPR) for INS prediction, and Gaussian regression process (GPR) for VO drift prediction. The operation process can also be divided into three stages: (1) train the monocular VO and INS drift predictor when the GNSS signal is available, (2) train the monocular VO drift predictor for modeling the error associated with the predicted speed of monocular VO. (3) In the case of GNSS signal loss, the monocular VO drift predictor is used for prediction. The main benefit of using this scheme is its ability to model different drift errors (monocular VO drift and INS drift) when GNSS measurements are available and to predict these errors during GNSS signal outages. In previously proposed regression algorithms based on GPR [66] and Support Vector Machine (SVM) [67], inconsistent matching can result from a lack of observed features or repeated patterns. The method proposed by Mostafa et al. effectively solves the serious problem of the inability to deal with missing optical flow vectors in some image parts caused by such inconsistent matching. Experiments show that the algorithm can effectively reduce positioning error during GNSS signal interruption. It reduces the positioning error to 47.6% and 76.3% of the pure VO/INS scheme and the VO/INS scheme with GPR correction, respectively, during a 1-minute GNSS signal outage. Xu et al. propose a multi-layer formulation for multi-target tracking that combines traditional optical flow constraints with product terms and uses a Sequential Convex Programming (SCP)-based approach to solve the resulting nonconvex optimization problem. To improve the accuracy and reliability of the autonomous navigation algorithm of the aircraft, they modeled and analyzed the errors of each sensor in the VO/IMU integrated navigation system, and studied the loose combination filtering equation based on Kalman filtering [68].

However, small systems such as UAVs have limitations in size, payload, and power, which are frequently encountered in the fields of computer vision and robotics. Because of the problem that the visual localization scheme requires high computing power and is difficult to miniaturize, in recent years, some researchers have also tried to improve the visual localization system. He et al. studied EBVO [69], a state-of-the-art edge-based VO, and proposed an optimization framework called PicoVO, which can greatly reduce computation and memory footprint. First, a lightweight edge detector is used to replace the Canny detector, the key processing stage of EBVO is optimized, and a processing scheme from sparse to dense is proposed in the tracking stage, instead of the image pyramid algorithm, and lightweight in the post-processing stage. This allows UAVs to use lightweight keyframe management in the post-processing stage. They evaluate a real RGB-D benchmark dataset on a NUCLEO-F767ZI equipped with a 216MHz Cortex-M7 MCU and 512KB RAM. PicoVO achieves a real-time frame rate of 33fps@320*240 on this platform, with high tracking accuracy comparable to state-of-the-art VO on PC [70]. Santamaria et al. [71] proposed a simple, low-cost, and high-rate state estimation method that enables MAVs to achieve autonomous flight with a low computational burden. The method uses a smart camera that integrates a monocular camera, an ultrasonic distance sensor, and a three-axis gyroscope, which can provide up to 200Hz optical flow measurement, range of reflective surfaces, and three-axis angular rate, avoiding the CPU overhead of real-time image processing. At the same time, the innovation of their proposed algorithm is that the linear velocity is not calculated from the optical flow information, and the motion state is directly observed using the original optical flow information, so the process and measurement noise are decouple.

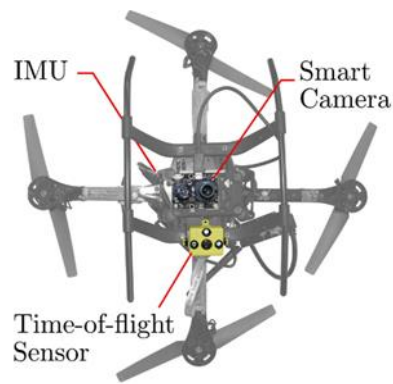

Fig.9 Tiny UAV with a smart camera underneath [29]

Pastor-Moreno et al. [29] designed OFLAAM for micro air vehicles (MAVs). The system consists of only a downward-pointing optical flow camera, a forward-pointing monocular camera, and an IMU. These features are mapped into a vocabulary through a localization module to address the loss of optical flow drift and localization estimation. This module uses the DBoW2 algorithm to perform position correction by loopback detection [72], and the combination of high-speed optical flow localization and low-rate positioning algorithm can realize fully autonomous localization of MAV while reducing the overall computational load [29]. Dong et al. [73] innovatively proposed a relative localization scheme for unmanned aerial vehicles. They acquired images of the ground through a camera mounted under the drone, then extracted feature points between two frames through the SURF algorithm, and gave the optical flow through the Fast Approximate Nearest Neighbors (FLANN) algorithm. The UAV speed can be calculated by the optical flow motion estimation equation and known parameters. Their proposed scheme integrates measurement data from SINS, electronic compass, optical flow, altimeter system, and laser rangefinder, giving relatively accurate localization information.
The table below gives a summary of the sensor fusion VO algorithms mentioned above, which lists the main problems and approximate solutions.

Table.3 Comparison of VO Algorithms for Sensor Fusion

| Method | Objectives | Solutions |
| --- | --- | --- |
| Shen et al.[28] | Implement multi-sensor fusion | Design a fusion algorithm based on EKF |
| Kim et al[63] | | Design a fusion algorithm based on FPTF |
| Yu et al.[64] | Solving scale ambiguity | Ground point aggregation algorithm to fit ground plane to estimate the scale |
| Mostafa et al.[27] | | Train scale drift predictors when GNSS signals are available |
| Xu et al[68] | | Error modeling analysis of each sensor |
| PicoVO[70] | Use navigation algorithms on low-computing platforms | Propose a performance optimization framework for EBVO |
| Santamaria et al.[71] | | Avoid high-performance requirements of real-time image processing with smart cameras |
| OFLAAM[29] | | Combining high-speed optical flow localization with low-speed positioning algorithms |
| Dong et al. [73] | | Directly Calculate Drone Velocity Using Optical Flow Motion Estimation Equations |

2.2.2 Speed/Distance/Position Estimation

How to estimate the motion state is an important topic in localization, and the VO algorithm based on optical flow/feature matching can provide rich self-motion information. Ho et al. used an extended Kalman filter (EKF) combined with captured images from a monocular camera to analyze the divergence of the temporal flow vector during the vertical landing of the acquired UAV, estimate the height and vertical velocity of the UAV and control the UAV landing [74]. For each captured image, they employed the FAST algorithm to detect corners and tracked them in the next frame using a Lucas-Kanade tracker, which allowed them to measure the contraction and expansion of optical flow and estimate the divergence of optical flow.

Using Optical Flow, Mumuni et al. used Structure of Motion (SfM) [75] as an addition to Monocular Depth Measurement (MDE) to improve depth estimation accuracy. In general, dense depth measurements derived from MDEs are ambiguous in scale, while sparse depth measurements provided by SfM models have a metric scale. Since MDE already provides dense depth measurements, only a few sparse depth information from SfM are needed to complement each other. Therefore, the combination of the two methods can improve the accuracy of the UAV's depth estimation. While CNN-based optical flow models have provided highly accurate estimation results in recent years [76, 77], the algorithm of Mumuni et al. is more memory- and computationally efficient [78]. To use optical flow for localization on small UAV platforms, a more efficient optical flow algorithm needs to be designed. Taking inspiration from a feature density distribution-based collision detection algorithm [79], McGuire et al. [80] propose the EdgeFlow algorithm, which introduces a variable time horizon to determine subpixel flow, and uses spatial edge distribution to track motion in images. Experiments verify that the method is computationally efficient enough to run at near-frame rates on limited embedded processors, providing reliable speed and distance estimates for UAVs in unknown environments. Similarly, since the integrated navigation system applied to indoor UAVs has a heavy demand for real-time optical flow calculation, Zheng et al. adopted the Lucas-Kanade sparse optical flow algorithm in their localization method and added Forward-Backward bidirectional tracking optimization. Considering the real-time performance, they use the ORB feature extraction algorithm, the feature matching uses the KNN forward-backward bidirectional method, and the matching results are filtered by the RANSAC algorithm. Experimental results show that this method has high velocity and position estimation accuracy [37]. In work by Huang et al. [81], a new end-to-end network is proposed for learning optical flow and estimating camera ego motion. They used the PWC-Net designed by Sun et al. [53] to estimate optical flow and adopted an autoencoder (CNN Encoder). A Recurrent Neural Network (RNN) is then used to examine optical flow changes and connections to compute the regression of the 6-dimensional pose vector. They used a deeper encoding network to learn effective optical flow features, improving the quality of the extracted optical flow space. In addition, they also train the encoder individually in an unsupervised manner, avoiding non-convergence in the entire network training process. They conducted extensive experiments on the KITTI and Malaga datasets, and the results show that the model has advantages in translation estimation and rotation estimation compared with other machine learning-based VO, and can also overcome the problem of scale ambiguity in traditional geometry-based VO.

2.2.3 Obstacle Detection

In general object detection, optical flow-based VO can also provide important information. The optical flow algorithm can know not only the position of the moving object, but also the speed and direction of the object. At the same time, the algorithm does not require background modeling and background update, so it is widely used [49]. There are currently three mainstream vision-based obstacle detection methods, including monocular cues, stereo vision, and motion parallax [82]. Among them, the motion parallax method is mainly based on optical flow, and the motion and structure of the observer and objects in the scene can be obtained from the optical flow field obtained from a series of images.

In the work by Meneses et al. [83], optical flow is directly used for obstacle detection instead of estimating the motion state of the robot. Based on the definition of optical flow, in the optical flow field, the relative motion of the area with low optical flow intensity is small, so the probability of containing obstacles is low. During robot motion, optical flow information is passed to an SVM classifier [84] with an RBF kernel, which indicates whether there are obstacles on the path to follow. The robot can update the movement direction based on this indication, usually the robot will be biased towards the direction with lower optical flow intensity [83]. Kendoul et al. adopted the dense optical flow method to provide information about the whole environment and used the Gunnar-Farneback method to estimate dense optical flow [40]. When the UAV moves forward, if a large number of optical flow vectors appear in the field of view, it indicates that there is an obstacle ahead. If the size of the optical flow vector is smaller than the threshold, it is considered that there is no object in front of the UAV. For the problem that foreground and background regions are mixed, they use different thresholds to extract regions according to the size of the optical flow vector. A clustering process is then performed to combine similar regions [82]. Zhang et al. [85]provide an auto-localization solution with tall buildings on both sides and no overhead view. They estimated its rotation by using the image obtained by the monocular camera and used the optimization algorithm to fit the points corresponding to the obstacles on both sides into a straight line using the RANSAC algorithm, which could minimize the error variance of the robot's forward-looking points. Then, the vanishing point of the straight line is calculated by the Kalman filtering algorithm, that is, the end where the parallel obstacles on both sides meet at infinity. The Ackermann steering model and singular value decomposition are used in the algorithm to estimate the direction of car motion, and wheel encoders are used to calculate the translation scale.

The reference of optical flow in CNN also facilitates the processing of moving objects in highly dynamic scenes. Rashed et al. improved semantic segmentation using motion information and depth information from optical flow for detecting roads, buildings, trees, etc. A CNN architecture based on semantic segmentation using multimodal information fusion, elaborated in the paper, focuses on autonomous driving applications, where prior information is leveraged to enhance the segmentation task [86].

# 3. Relative Visual Localization (VSLAM-based)

However, monocular VO-based methods cannot directly obtain precise distances, which may limit their use in some specific tasks. In contrast, SLAM-based methods can provide precise metric maps through sophisticated SLAM algorithms. VO often forgets the perceived world structure, while VSLAM keeps a long-term map of the world. VSLAM uses one or more cameras as the main perception sensor, which is generally more economical than Lidar-based SLAM systems, while also providing superior localization performance on multiple datasets [87]. In the absence of absolute position signals such as GPS, VO/VSLAM can be used as an effective supplement to the wheel odometer/inertial odometer. Visual-inertial SLAM (VI-SLAM) uses two complementary data streams to achieve better accuracy and robustness as well as a higher frequency of state estimation compared to VO/VSLAM which only relies on visual sensors. A complete visual SLAM framework consists of sensor data, front-end visual odometry, backend nonlinear optimization, loop detection, and mapping [31].

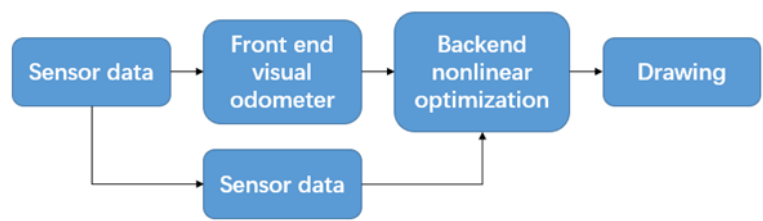

Fig.10 Schematic diagram of the visual SLAM framework

VO and IMU fusion is a commonly integrated navigation algorithm in VSLAM, which is essentially a combination of speed, that is, the optical flow localization system and the inertial navigation system work independently. The difference between the speed information obtained by the optical localization algorithm and the localization information calculated by the inertial navigation system can be used as system input, and the speed error of the inertial navigation system can be estimated through the Kalman filter, thereby correcting the error of the inertial navigation system. According to whether the image feature information is added to the state vector, it can be divided into two categories: loosely coupled and tightly coupled [88]. Loose coupling estimates the states of the visual localization system and the IMU system separately to calculate the estimation results; tight coupling is to estimate the camera pose and the state measured by the IMU as a whole. Usually, the performance of tight coupling is better.

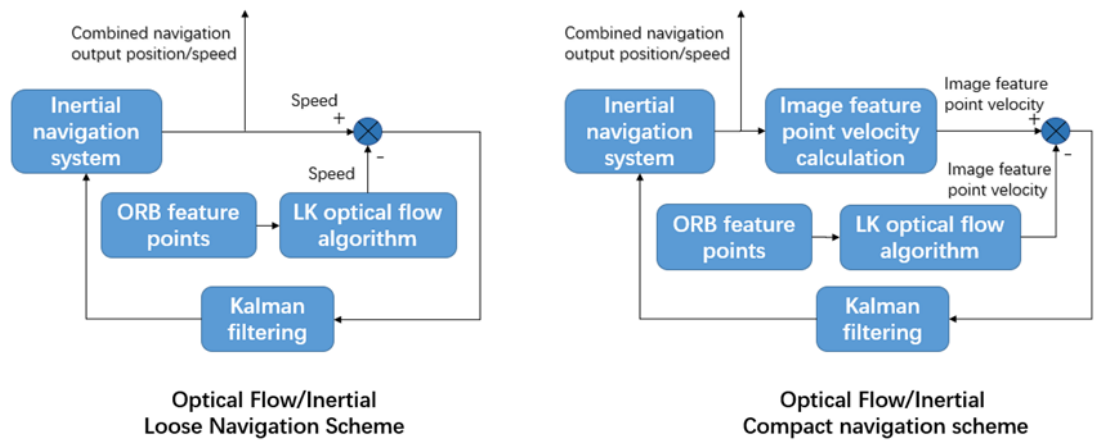

Fig.11 Schematic diagram of loose and tight coupling in an integrated navigation scheme

State-of-the-art combined VO/IMU solutions (based on tight coupling) can be broadly classified as either optimization-based or extended Kalman filter-based systems. Optimization-based solutions including
- ORB-SLAM [3] is a monocular SLAM system based entirely on sparse feature points, employing ORB as the core feature. It implements a complex sensor fusion procedure and adds the detection and closing mechanism of the loop so that it can estimate the state and error more accurately. The algorithm can rectify the constructed sparse feature map while allowing the camera pose to be relocated in the sparse feature map without any prior location information. ORB-SLAM2 [4] adds support for binocular cameras and RGB-D cameras based on ORB-SLAM. It uses the Bundle adjustment method to obtain the camera pose and coordinates while obtaining the reconstruction of the surrounding objects. ORB-SLAM3 [5] implements feature-based tight coupling and only relies on maximum a posteriori estimation, which effectively improves the localization accuracy in different scenarios. The algorithm also introduces the Atlas system. When the tracking process is lost, the system can re-align with the originally constructed map according to the current frame, which enhances the robustness of localization.
- OKVIS [6] is based on a tightly coupled VIO algorithm that employs Harris corner and BRISK feature point detection for temporal and stereo matching, uses all visual measurements of relevant keyframes to estimate motion states, predicts state quantities by using IMU sensors, and optimizes A cost function consisting of parameters to perform nonlinear optimization.
- VINS-Mono [7] implements tightly coupled visual and inertial joint state estimation,

which treats state estimation as a nonlinear least squares problem and optimizes over a moving window of data measurements. The algorithm can calculate the sensor's external parameters and time delay while ensuring the effect of high-precision VO.

In contrast, solutions based on extended Kalman filters include:
- ROVIO [8] completed the data fusion of vision and inertial sensors using an iterative Kalman filter algorithm. It uses QR decomposition to reduce the computational complexity of the least squares problem by simultaneously tracking the features of multiple multi-level image blocks and accelerating the solution speed.
- MSCKF [9] is based on a visual-inertial algorithm with an extended Kalman filter, using IMU measurements to predict the filter state and visual feature measurements to update the state vector. Compared with the general VO algorithm, the algorithm is more robust, can adapt to the environment with sparse visual texture, and is more suitable for running on a platform with limited computing resources.
- SVO [10, 11] is visual odometry that combines the feature point method and the direct method. Compared with ORB-SLAM, its matching method is changed from the feature-based method to the gray value matching method. Foster et al. improved the SVO algorithm and proposed SVO_PRO. The algorithm has good support for monocular/binocular/array cameras. At the same time, it uses OKVIS to optimize the pose and feature points, which further improves the algorithm performance.

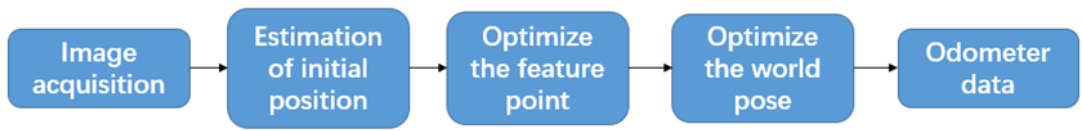

Fig.12 Schematic diagram of the SVO algorithm

The EKF-based solution models the state estimate as a normal distribution, linearizes the state equation, and applies the EKF to the resulting error coordinates. While feature-based methods usually achieve the highest accuracy in computing robot trajectories, EKF-based methods still receive attention due to their lower memory requirements and processing time [89].

The above VSLAM method has been widely used in recent years, and the following is a brief introduction to the applications of this method.
Cheng et al. [90] improved the original ORB-SLAM by using optical flow to distinguish and eliminate the extracted dynamic features, avoiding the impact of dynamic objects on VSLAM performance. Zheng et al. [91] made improvements to the previous VSLAM method on point features and tried to incorporate line features into the calculation to avoid the lack of robustness of point features in environments with low texture and illumination changes. They propose a novel VSLAM method based on tightly coupled filtering named Trifo-VIO, in which the line features are used to help improve the robustness of the system in challenging scenarios. Based on previous work on line features, Lim et al. [92] proposed a method to identify degenerate lines for the degradation problem that occurs when line features are used. Furthermore, to improve the robustness of line matching, they propose an optical flow-based line tracking method. Experiments on the EuRoC dataset validated the results of the improved monocular

SLAM system, which proved to produce more accurate localization and mapping results than VINS-Mono. Li et al. [93]carried out similar work and proposed a midpoint-based fast line feature description and matching method, and compared the line segment detector (LSD), fast line detector (FLD), and edge drawn lines (EDLines) three Line detection algorithm. Furthermore, they combined the above method with VINS-Mono and proposed FPL-VIO. Compared with similar line feature-based algorithms, FPL-VIO improves line processing efficiency by 3-4 times while maintaining the same accuracy [93]. The improvement of the line features over VSLAM methods can also be seen in the work [94–97].

Lin et al. proposed a multi-sensor fusion method based on ORB-SLAM, IMU, and wheel odometer. IMU and wheel odometer provide relatively accurate measurements of robot movement distance in the world coordinate system, to correct the problem of mapping scale deviation that may exist in ORB-SLAM during localization. Experiments show that it has strong localization ability in GPS-denied scenes [98]. Shan et al. proposed the VINS-RGBD system, which is built on VINS-Mono and fuses IMU and RGBD camera data to utilize depth data during the initialization process as well as the VIO stage. Furthermore, they integrated a mapping system based on subsampled depth data and octree filtering in real-time mapping. The VINS-RGBD method outperforms VINS-Mono and ORB-SLAM in evaluation experiments on hand-held, wheeled, and tracked robots [99]. Similarly, Xu et al. [100] designed a Real-Time Locating System (RTLS) that can be applied in a GPS-denied environment. In this system, the additional Occupancy Grid Map is jointly constructed with the sparse feature map of ORB2-SLAM (RGBD camera), which can also use less memory in large-scale environments, and achieves user interaction and path planning. Their proposed RTLS system and RTLS do not require any environmental instrumentation or rely on any existing human infrastructure, which makes its rapid deployment possible.

Considering that the stereo vision system is more robust in different environments and motion situations, Sun et al. [101] chose the MSCKF algorithm as the starting point and designed the stereo VIO solution S-MSCKF . Contrary to popular belief that stereo vision-based estimation yields higher computational cost than monocular methods, their proposed stereo VIO has similar or even higher efficiency than OKVIS and VINS-MONO. Similar work has been done by [101] and [102, 103] . Based on MSCKF, Zhang et al. used a variational Bayesian adaptive algorithm to further improve the performance of the algorithm. They modeled the noise as observation noise. The covariance was modeled as an inverse Wishart distribution with harmonic mean rather than Gaussian noise with constant mean and covariance assumed in the original algorithm [104]. Patrick et al. also significantly improved the algorithm performance by adjusting the Schmidt-Kalman formulation within the MSCKF framework [105]. Similarly, to address the problem that under special motions (such as local acceleration), VIO will be affected by additional unobservable directions, resulting in larger localization errors, Ma et al. designed ACK-MSCKF. The algorithm fuses the Ackermann error state measurement with the tightly coupled filter-based mechanism in S-MSCKF, exploiting the additional constraints of the Ackermann measurement to improve the pose estimation accuracy. ACK-MSCKF significantly improves the pose estimation accuracy of S-MSCKF under special motions of autonomous vehicles and maintains accurate and robust pose estimation under different vehicle driving cycles and environmental conditions [106]. Other recent studies on MSCKF include [107–109].

The following table is a summary of the above VSLAM method.

Table.4 A summary of the VSLAM approach

| Method | Prototype | Feature |
| --- | --- | --- |
| Cheng et al.[90] | ORB-SLAM | Using Optical Flow to Distinguish and Eliminate Extracting Dynamic Feature Points |
| Lin et al.[98] | ORB-SLAM | Correcting the bias of the VSLAM algorithm using the IMU |
| Xu et al.[100] | ORB-SLAM | Implementing the joint construction of OGM and sparse feature map |
| FPL-VIO[93] | VINS-Mono | Using line features to improve monocular SLAM effects |
| VINS-RGBD[99] | VINS-Mono | Fusing IMU and RGBD camera data in VSLAM |
| S-MSCKF[101] | MSCKF | Applying the Binocular Camera to MSCKF |
| Zhang et al.[104] | MSCKF | Improving Performance Using Variational Bayesian Adaptive Algorithms |
| Patrick et al.[105] | MSCKF | Adjusting the Schmidt-Kalman formula within the framework to improve performance |
| ACK-MSCKF[106] | MSCKF | Using additional constraints from Ackermann measurements to improve pose estimation accuracy |

## 4. Absolute Visual Localization(image registration-based)

The main advantage of AVL over RVL is that it is not affected by drift. The algorithm typically uses offline map data to achieve positioning, providing accurate, drift-free localization even in the absence of external positioning signals. Currently, AVL, which relies on offline map registration, has become a popular method for localization and navigation in GPS-constrained environments. Offline map data is mainly derived from orthorectified satellite imagery [12] and assumes that the image has been precisely matched to the actual location. The main goal of AVL is to match the UAV's currently acquired imagery with offline map information to perform localization, achieving immunity to drift by ensuring complete independence between estimates [110]. However, algorithms based on offline map registration are still very vulnerable to seasonal landscape effects. In the work by Fragoso et al., they try to solve this problem in the domain of image transformation architecture using deep learning methods to transform seasonal images into stable and invariant. The transformed images can be used by traditional algorithms without modification, so the geometry and uncertainty estimates of traditional methods are preserved, and they still show excellent performance under extreme seasonal variation, while being easy to train and highly generalizable. Experiments show that their proposed algorithm almost eliminates severe mismatches in practical visual localization tasks, which also include topographic and perspective effects [13]. Shan et al. proposed a method to assist UAV position localization with Google Maps. First, the position of the drone is predicted by optical flow. During attitude tracking, the histogram of oriented gradients (HOG) features and particle filters are used to match the ground imagery acquired by the drone with the geographic imagery recorded by Google Maps to locate the drone. This algorithm avoids the relatively inefficient sliding window search [14], while the search area is limited to the location near the optical flow prediction, which greatly improves the search efficiency [15].

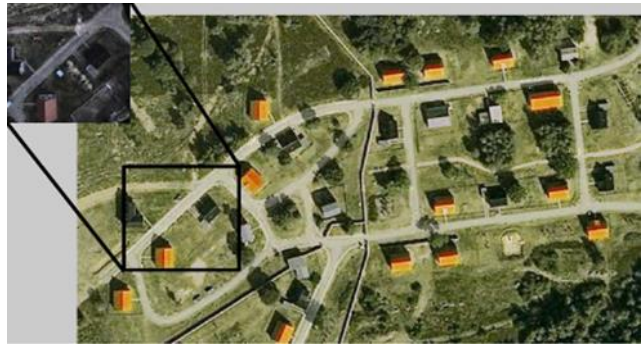

(a) The image obtained by UAV and the upper left corner is the image of the target area.

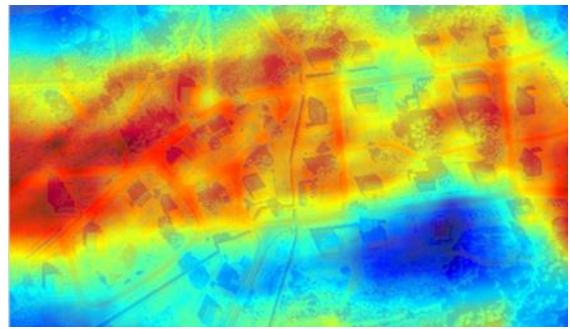

(b) Confidence map of the match. Red indicates high confidence, while blue indicates low confidence.

Fig.13 Image matching using HOG features and particle filters [14]

Dumble introduced a vision-aided inertial navigation system that uses ground features (in this case, road intersections) matched to a database to provide position measurements. The shapes of road intersections are extracted from visual images using RANSAC and then matched against a reference database to provide image-to-mapped road intersection correspondences. The algorithm allows the drone to locate itself in GPS-denied areas [16].

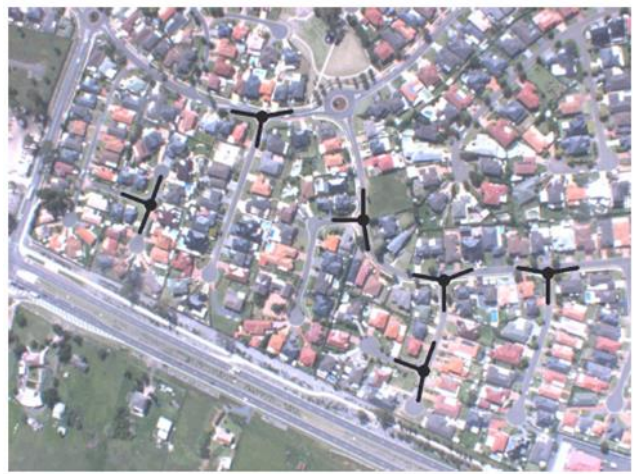

Fig.14 Fitting road intersections using RANSAC [16]

Another typical approach to AVL combines information from known environmental features such as roads and lanes, and the algorithm achieves good performance in GPS-denied environments without road restrictions. Line markings on the lane floor (such as double yellow, broken yellow, and broken white) often correspond to a set of GPS waypoints for each feasible lane of the map information. These line markings are critical to vehicle navigation, and ignoring this information can often lead to catastrophic behavior, such as autonomous localization systems steering the vehicle down the wrong road or even sidewalks. The problem is compounded by the fact that unspecified markers on the map (such as unmeasured lane lines, crosswalks, or turn arrows) may also exist. Miller et al. proposed a vision-aided localization system called PosteriorPose [17] that uses vision, IMU, and Bayesian particle filters and map information to aid localization [18]. The algorithm makes vision-based measurements of nearby lanes and stop lines based on a map of known environmental features. These map-related measurements improve the quality of the localization solution when GPS is available and can keep the localization solution converged during prolonged GPS outages. Gruyer et al. have also done similar work. The figure below shows the ground images they obtained through cameras mounted on both sides of the car, and the lane types and measurement results extracted from the images [23].

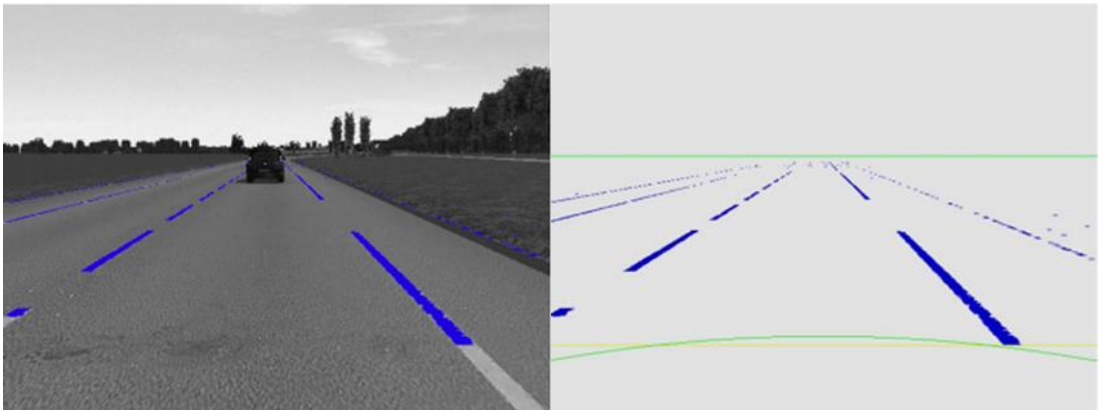

Fig.15 Lane marking primitive detection using the front camera. Left: The ground image obtained by the vehicle camera; Right: The detected and extracted lane lines [17]

The algorithm outperforms a tightly coupled GPS/Inertial Localization solution in the event of prolonged GPS signal loss. A similar approach was adopted in the work of Tao et al., but their goal was to reduce the localization error of GNSS [19]. Occupy grid map (OGM) has been used for map-assisted localization. In recent years, in response to the known shortcomings of OGM, such as the discretization of the environment, the assumption of independence between grid cells, and the need for measurement accuracy, researchers developed Gaussian Process Occupation Maps (GPOMs) [20]. Hata et al. [111] designed a new likelihood model based on a multivariate normal probability density function and employed a particle filter localization approach to work with GPOM. Experiments show that the localization error is more than three times lower compared to particle filter localization using OGM. In another study by Hata et al., they designed a novel vehicle localization technique for urban environments based on GPOM. The method utilizes roadside and road marking data to construct GPOM and OGM. For the case of sensor noise and low confidence, the localization method is based on the Monte Carlo

Localization (MCL) algorithm [21]. Similarly, Yuan et al. improved the framework of GPOM to reduce cost while maintaining good performance, allow GPOM to run online, and obtain relatively better quality than classical GPOM [22]. Other work on assisted localization based on lane detection can also be found in [24, 25].

# 5. Future research

Current research has shown that visual methods can be used to provide useful information in GPS-denied environments. Researchers are conducting research in the area of improving the accuracy, robustness, reliability, and transferability of visual navigation methods. In the field of vision-based navigation methods in GPS-denied environments, there are still some problems to be solved.

**1. Miniaturization and cost of the vision system**

First, the vision system already saves hardware costs in the information acquisition process compared to radar-based systems. However, since the information obtained by the vision system is relatively rich (including brightness information, depth information, color information, etc.), a large number of computing resources are required to process the obtained information, resulting in the need for the support of strong computing hardware. This also makes it difficult for visual localization systems to be mounted on small platforms (such as autonomous micro-UAVs), or to realize real-time processing of information on small platforms. The algorithm based on optical flow is an effective way to solve this problem. Optical flow-based methods are more efficient because the algorithm does not require complex feature point descriptors and can track valid features at the original intensities of the image, recovering most of the motion information available in the image. The trade-off between sparse and dense features also allows the algorithm to strike a balance between performance and quality. Traditional image processing optimization schemes (such as image pyramids) are also widely used in the algorithm. In addition, in platform applications that require multi-sensor fusion, the extended Kalman filter algorithm has always received continuous attention due to its lower memory requirements and processing time. In the studies mentioned above, we found that some researchers are working towards miniaturization, but more research is still needed.

**2. Robustness of the visual system**

Low reliability in navigation systems has the potential to be catastrophic. Due to common reasons such as lens distortion, lack of texture, and insufficient light, the information directly obtained by the visual system may be quite different from the actual one, so the measuring scale (such as the estimated distance to an obstacle) is likely to be unreliable. Hardware upgrades (such as the use of RGB-D cameras) may be an effective measure. In addition, for the problem of scale ambiguity, accurate and robust scale recovery algorithms are of great significance in applications. Scale recovery is to correct the measurement scale in combination with absolute reference information. Absolute reference information can come from the IMU, wheel encoders, or stereo cameras. However, these sensors are not always available in practical applications, and efficient sensor fusion algorithms are required to achieve scale correction. In addition, in the method mentioned above, the distance between the camera and the ground is obtained by extracting the feature points on the ground and fitting the ground plane directly, which is also a promising solution for the scale restoration task based solely on visual information. Deep learning methods are also widely used in error correction, trying to solve problems by training CNNs on large amounts of data. For example, we can try to correct the problem of scale ambiguity by introducing stereo image training or an additional scale-consistent loss term in the network. Due to the limited availability of labeled data, unsupervised deep learning methods are also becoming a research hotspot in this direction. However, although deep learning methods have achieved advanced results in error correction, they are still inferior to some traditional methods in terms of generalization, efficiency, and accuracy. Finally, VSLAM closed-loop and map registration in absolute localization are also effective correction measures. The VSLAM closed-loop achieves location calibration by identifying pre-visited regions, while the map alignment criterion accomplishes this task based on offline/online maps. With the abundance of map information (street view maps, satellite maps) resources, we believe that the work on map registration will shine in the future.

**3. The practical feasibility of the vision system**

Currently, the datasets widely used in visual localization algorithms include Middlebury, KITTI, Sintel, etc. In the optical flow algorithm application scenario, the Flying Chairs dataset is also widely used. Many new VO algorithms show the feasibility and potential on many datasets, but there is a shortage of VO methods that can be applied to the real world. Therefore, it is also a challenging task to apply the algorithm to practical scenarios in the real world. But in recent work, we have also found a wide range of applications of visual localization algorithms in agricultural production, autonomous UAV flight, indoor navigation, and other directions. We hope to see more visual methods put into practical use in the future.

# 6. Conclusion

This paper reviews vision-based localization methods in GPS-denied environments, including optical flow (feature extraction)-based visual odometry methods, visual odometry methods based on SLAM, offline map registration, and lane matching-based methods, etc. We classify research related to visual localization algorithms into traditional methods (feature extraction-based) and unconventional methods (deep learning-based methods). We introduce key working principles for each category and, where applicable, describe applications associated with each method.

By reviewing the related work on visual localization methods in recent years, we believe that solutions are developing towards low cost, miniaturization, and high precision. With the rapid development of deep learning in recent years, some researchers have tried to use neural networks to solve visual localization problems and have achieved some results. The most distinctive one is FlowNet based on optical flow extraction and its improved algorithm, which has made important contributions to extracting image motion information and providing localization support. At the same time, with the development of various hardware sensors (IMU, wheel encoder, etc.), vision technology is being fused with multi-sensors, using the advantages of various sensors to complement each other to achieve reliable accuracy and robustness. In addition, with the enrichment of map databases, visual localization methods based on map registration are also becoming a promising research direction. As a long-term researched area, lane detection also shows potential in applications.

The expected visual localization system should have environmental awareness, can effectively handle outliers, adapt to environmental challenges, and provide reliable, robust, and accurate state estimation in real-time, enabling autonomous localization under GPS-denied conditions. According to the surveyed papers, major future research directions include how to achieve robust and reliable pose estimation and localization planning in GPS-denied, complex, and visually degraded environments.

Bayesian adaptive nonlinear filter," *IEEE Sensors Journal*, vol. 20, no. 16, pp. 9437–9448, 2020.